\documentclass[conference]{IEEEtran}
\IEEEoverridecommandlockouts
\usepackage{cite}
\usepackage{amsmath,amssymb,amsfonts}
\usepackage{algorithmic}
\usepackage{graphicx}
\usepackage{textcomp}
\usepackage{booktabs}
\usepackage{multirow}
\usepackage{xcolor}
\usepackage[colorlinks,linkcolor=blue,anchorcolor=blue,citecolor=blue]{hyperref}
\makeatletter

\newcommand{\Rmnum}[1]{\expandafter\@slowromancap\romannumeral #1@}
\makeatother

\def\BibTeX{{\rm B\kern-.05em{\sc i\kern-.025em b}\kern-.08em
    T\kern-.1667em\lower.7ex\hbox{E}\kern-.125emX}}
\begin{document}

\title{A Deep Learning Framework for Trafﬁc Data Imputation Considering Spatiotemporal Dependencies\\
\thanks{This research is funded by the Shenzhen Science and Technology Innovation Commission (Grant No. JCYJ20210324135011030, WDZC20200818121348001 and KCXFZ202002011010487), the National Natural Science Foundation of China (Grant No. 71971127), Guangdong Pearl River Plan (2019QN01X890), and the Hylink Digital Solutions Co., Ltd. (120500002).}
}

\author{
\IEEEauthorblockN{Li Jiang}
\IEEEauthorblockA{\textit{Tsinghua-Berkeley Shenzhen Institute} \\
\textit{Tsinghua University}\\
Shenzhen, China \\
 jl20@mails.tsinghua.edu.cn}
\and
\IEEEauthorblockN{Ting Zhang}
\IEEEauthorblockA{\textit{Tsinghua-Berkeley Shenzhen Institute} \\
\textit{Tsinghua University}\\
Shenzhen, China \\
 zhangt20@mails.tsinghua.edu.cn}
\and
\IEEEauthorblockN{Qiruyi Zuo}
\IEEEauthorblockA{\textit{Tsinghua-Berkeley Shenzhen Institute} \\
\textit{Tsinghua University}\\
Shenzhen, China \\
 zqry20@mails.tsinghua.edu.cn}
\and
\IEEEauthorblockN{Chenyu Tian}
\IEEEauthorblockA{\textit{Alibaba Group} \\
\textit{Tsinghua University}\\
Hangzhou, China \\
 tianchenyu.tcy@alibaba-Inc.com}
\and
\IEEEauthorblockN{George P. Chan}
\IEEEauthorblockA{\textit{Loudonville Christian School} \\
Loudonville, NY\\
gchan@lcscourses.net}
\and
\IEEEauthorblockN{Wai Kin (Victor) Chan$^*$}
\IEEEauthorblockA{\textit{Tsinghua-Berkeley Shenzhen Institute} \\
\textit{Tsinghua University}\\
Shenzhen, China \\
 chanw@sz.tsinghua.edu.cn}}

\maketitle

\begin{abstract}
Spatiotemporal (ST) data collected by sensors can be represented as multi-variate time series, which is a sequence of data points listed in an order of time. Despite the vast amount of useful information, the ST data usually suffers from the issue of missing or incomplete data, which also limits its applications. Imputation is one viable solution and often used to prepossess the data for further applications. However, in practice, n practice, spatiotemporal data imputation is quite difficult due to the complexity of spatiotemporal dependencies with dynamic changes in the traffic network and is a crucial prepossessing task for further applications. Existing approaches mostly only capture the temporal dependencies in time series or static spatial dependencies. They fail to directly model the spatiotemporal dependencies, and the representation ability of the models is relatively limited.
To better capture the complex spatial-temporal dependencies and impute data, we propose a new ST data imputation model. Temporal convolution and self-attention networks are utilized to capture long-term dependencies and dynamic spatial dependencies, respectively. Furthermore, our model develops a novel self-learning node embeddings to learn the intrinsic attributes of different sensors.An end-to-end framework incorporates these elements. We empirically illustrate the benefit of our proposed framework by comparing other algorithms in real-world data sets.
\end{abstract}

\section{Introduction and related work}
Numerous disciplines, including criminology, social science, and transportation, frequently encounter spatial-temporal features.

With the recent proliferation of information and communication technology (ICT), the scale and dimension of spatiotemporal (ST) data in transportation areas have become a key input for a wide range of applications, such as urban planning, resource allocation, and keypoint identification.

In intelligent traffic systems (ITS), data collection is a fundamental module. It is common to see various kinds of detectors, such as loop detectors, radars, and cameras on roads, which are used to record traffic information such as speed, volume, and traffic density. Urban computing studies, which draw from the fields of wireless and sensor networks, information science, and human-computer interaction, are considerably aided by big data research. Data collection on urban transportation environments aids in raising the standard of living for those who live in metropolitan areas. Urban computing is an interdisciplinary field that also has practitioners and applications in areas including energy, urban planning, ethnography, public history, health care, and civil engineering, among others.

A few researchers have reviewed recent literature on the problems of ST data applications in different fields. Compared with mining patterns from traditional relational data, modeling ST data is particularly challenging due to its spatial and temporal attributes in addition to the actual measurements. The fact that time and space exist introduces a wide range of ST data types, which gives rise to various ST data mining issues and methodologies. For modeling ST data, Atluri et al. \cite{atluri2018spatio} have evaluated the common issues and approaches. To pinpoint the pertinent issues for any form of ST data in practical applications, a taxonomy of the various ST data types and means of identifying and expressing data instances has been presented.  They also listed the commonly studied ST problems and reviewed the issues for dealing with unique properties of different ST types. Wang et al. \cite{wang2020deep} examined current developments in applying deep learning techniques to ST data mining applications and suggested a pipeline for the utilization of deep learning models for ST data modeling problems. Different from other types of data, ST data collected by sensor networks has some properties.
\begin{enumerate}
    \item \textbf{Temporal dependence}. There are nonlinear temporal dependencies. For example, the traffic situations can change  dynamically, regularly, and periodically, such as during the morning rush hour and evening rush hour in a day), which in turn impacts the correlations between diverse time steps.
    \item \textbf{Spatial dependence}. There are dynamic spatial connections on complex networks, which means the dependencies of nodes in the network of roads can vary over time if we take different traffic situations into account. For instance, traffic congestion negatively influences the upstream while having little influence on the downstream. 
    \item \textbf{Auto-correlation}. There are typically geographical and temporal connections between nearby measurements among the ST data observations, which make them not entirely independent. 
\end{enumerate}

The issues with missing data persist despite technological advancements and are difficult to resolve. In Orlando, Florida, data acquired by loop detectors on I-4 had a 15\% missing rate, according to Chandra et al\cite{al2004new}. According to an empirical assessment, the Georgia NaviGAtor system's ST data collection had an average missed rate of between 4\% and 14\%\cite{ni2005markov}. In Beijing, China, the daily traffic volume data had an average missing rate of 10\% (4\% from detector malfunctions and 6\% from other causes), with certain detectors having a missing rate as high as 25\% \cite{5169998}.

The machine learning methods are popular and are quite accurate and efficient in ST studies. 
For example, Lee et al. \cite{han2016spatiotemporal} presented a K-nearest neighborhood (KNN) based ST data imputation approach which is able to utilize the spatiotemporal information underlying traffic data. Chang et al. \cite{6216760} proposed a refined KNN method with Local Least Squares to impute the incomplete data.  Tak et al. \cite{tak2016data} did data imputation in sectional units of road links via transforming raw data into several two-dimensional images using KNN, which made use of both temporal and spatial data. Compared with traditional machine learning methods, deep learning methods have stronger model representation ability and achieved superior performance in many areas such as computer vision, natural language processing, and ST data mining. The success also draws the attention of researchers from other related areas. For instance, Li et al. \cite{LI2020105592} proposed a multimodal deep learning model to enable heterogeneous traffic data imputation. The model involves the use of two parallel stacked autoencoders that can simultaneously consider the spatial and temporal dependencies, Asadi and Regan \cite{Asadi2019} proposed a convolution bidirectional-LSTM for capturing spatial and temporal patterns and analyzed an autoencoder’s latent feature representation.  Zhuang et al. \cite{zhuang2018innovative} reformed the imputation problem into an image reconstruction pr oblem which focuses on making up the missing regions by introducing the convolution neural networks(CNN), which outperformed others in the high missing situation. Initially, Generative Adversarial Networks (GANs) were proposed as an idea for semi-supervised and unsupervised learning by Ian Goodfellow \cite{goodfellow2014generative}. 

The goal of this study is to build a deep learning model for ST data imputation by using the deep learning framework considering ST dependencies inherently existing in the data. In Section \Rmnum{2}, we give a brief introduction of the data set we have used. In section \Rmnum{3}, we formalize the ST data imputation problem and introduce the overall framework. In section \Rmnum{4} ,experiments are implemented on real-world data sets and we compare the results with other models. Then we explain the details of the model in Section \Rmnum{5}. Finally, we summarize our work results and conclude future research directions.

\section{Data sets}
We used three data sets in the context of public transportation. The ﬁrst is of trafﬁc speed in Guangzhou (GZ), China, consisting of 214 anonymous road segments measured within two months from August 1, 2016, to September 30, 2016 (61 days) at a 10-minute interval (\textit{GZ Data}). The second is of passenger ﬂow in Hangzhou (HZ), China, consisting of incoming passenger ﬂow of 80 metro stations over 25 days from January 1 to January 25, 2019, with a 10-minute interval (\textit{HZ Data}). The last data set is of trafﬁc speed in California, USA, consisting of 325 sensors in the Bay Area lasting for six months ranging from January 1 to June 30, 2017,with a total amount of 181 days. The total number of observed trafﬁc data is 16,941,600 and the time interval is 5-minute (\textit{PeMS Data}) and is provided by the California Department of Transportation (\textit{Caltrans}).

The data sets are diverse in the number of roads, days, and time intervals for measurements, as seen on Table \ref{tabel1}.
\begin{table}[htbp]
\centering
\caption{Description of the data set.}
\begin{tabular}{|c|c|c|c|} 
\hline
Dataset~ & \#roads & \#days & interval  \\ 
\hline
GZ & 214     & 61     & 10 min    \\
HZ  & 80      & 25     & 10 min    \\
PeMS      & 325     & 181    & 5 min     \\
\hline
\end{tabular}
\label{tabel1}
\end{table}

\section{Methods}
In this section, we will first formally define the ST data imputation problem from the mathematical view and then introduce the picture of the whole structure and the details of its building blocks: the temporal convolution layer and the dynamic attention layer.

\subsection{Problem setup} 
$\mathrm{X}(\mathrm{t}) \in R^{N}$ means the observation values of traffic at time step $t$, where $N$ is the quantity of detectors in the network. For a finite time steps $T$, we have a dataset of historical observation $\mathcal{X} = \left(X^{1}, \cdots, X^{T}\right)$, and $\mathcal{X} \in R^{T * N} $ is set of observations from time step $0$ to $T$. Then, we define the mask $\mathcal{M} = \left(M^{1}, \cdots, M^{T}\right)$, which has the same shape as $\mathcal{X}$. The elements in $\mathcal{M}$ take value in $(0, 1)$. We will call $\mathcal{X}$ the data matrix, $\mathcal{M}$ the mask matrix.

$\mathcal{M}$ indicates which observations of $\mathcal{X}$ are observed. In other words, if $\mathcal{M}_{i,j}=0$, the corresponding data in the data matrix is missing. The missing is random at given missing rates.
In the imputation setting, the objective is to impute the unobserved values with reasonable ones based on partially observed data and return a complete dataset.
In practice, to impute the data $X^{t}$, observations before and after time $t$ are used as inputs. Without adding any complexity, we define
\begin{equation}
\begin{aligned}
        & \mathcal{X}_p = (X^{t-T_{p}+1}, \cdots, X^{t-1})\\
        & \mathcal{X}_f = (X^{t+1}, \cdots, X^{t+T_{f}}) \\
        & \mathcal{M}_p = (M^{t-T_{p}+1}, \cdots, M^{t-1})\\
        & \mathcal{M}_f = (M^{t+1}, \cdots, M^{t+T_{f}}),
\end{aligned}
\end{equation}
where $T_{p}$ and $T_{f}$ are past time steps and future time steps, respectively.
Then the learned function $f(\cdot)$ that map past and future observations to current observation at time $t$ as
\begin{equation}
    \left[\mathcal{X}_p, \mathcal{X}_f, \mathcal{M}_p, \mathcal{M}_f \right] \stackrel{f(\cdot)}{\longrightarrow} \hat{X^{t}}.
\end{equation}

We highlight that our problem setup is different from ST data prediction tasks \cite{afrin2022long, yan2022short, tian2021spatial} that most researchers are focused on. The biggest difference between our problem setup with prediction tasks is available for the future dataset $\mathcal{X}_f$ while prediction tasks are not. It shows the potential that ST data imputation problem setup can use the future information to fill the missing data, and then better performances may be achieved in ST data prediction tasks. 

\subsection{Framework of the STAWnet}
\begin{figure}[htbp]
\centering
\includegraphics[width=.3\textwidth]{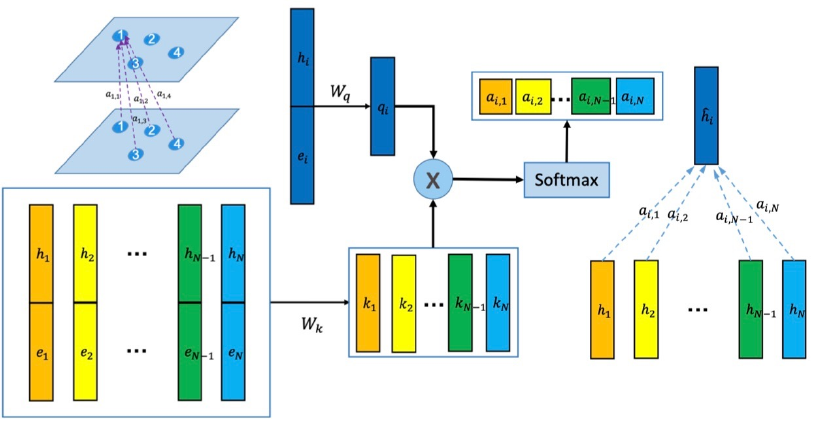}
\caption{There are various ST-blocks in the STAWnet. A gated TCN and a DAN are included in every ST-block, and node embedding is included into both of them. In order to prevent over-fitting, network layer normalization\cite{ba2016layer} is used inside each block. Additionally, the network makes use of residual and skip connections to hasten convergence and gather more detailed data. The skip outputs from the gated TCN in the various ST-blocks are then added together. Finally, the sum computes the predictions of the missing ST data via output layers.}
\label{stawnet}
\end{figure}

We borrow the idea from \cite{tian2021spatial} and adjust its framework STAWnets in our ST data imputation setting, shown in Fig \ref{stawnet}. We little abused the framework name by following its name from \cite{tian2021spatial}. Multiple stacked spatial-temporal blocks (ST) and output neural network layers make up the STAWnet's structural structure. It comprises of a dynamic attention network (DAN) and a gated temporal convolution network (TCN) for each ST-block. The goal of TCN and DAN is to obtain the temporal and spatial information and dependencies, respectively. After simply stacking these ST blocks, it can find the ST information and dependencies at different temporal levels. We show elements of each modular in the following.


\subsection{Gated TCN for extracting temporal information}
\begin{figure}[htbp]
\centering
\includegraphics[width=0.2\textwidth]{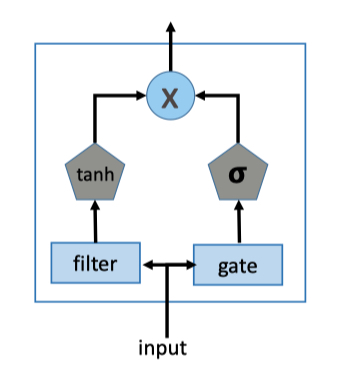}
\caption{Framework of Gated TCN}
\label{tcn}
\end{figure}

RNN-based methods are prevailing in time-series prediction and analysis.However, these methods have the disadvantages of excessive time-consuming and gradient explosion or vanishing in practice. On the other hand, CNN-based methods benefit from parallel processing, steady gradients, and simple structures, making them stable and time-efficient.

Inspired by \cite{van2016conditional, wavenet, tian2021spatial}, we follow the structure of \cite{tian2021spatial}. In contrast to only focusing on ST data prediction tasks, we leverage the STAWnet framework for ST data imputation tasks. Following their framework, we utilize the dilated CNN that enjoys the advantages of capturing an exponentially large receptive fields by the increase of the neural networks' depth.Dilated convolution networks involve filters that are applied in the area larger than its length by skipping the input values within a certain step. It bears some similarities to traditional convolution neural networks with a larger filter derived from the original architecture by dilated convolution with masks. The convolution networks become more efficient and effective with a larger filter. For the special example, dilated convolution networks with dilation 1 result in the traditional convolution networks. In addition to dilated convolution neural networks, gated mechanisms can help our whole framework to learn complex temporal information and dependencies. Gated TCN is written as

\begin{equation}
    \mathcal{X}_{T}^{l}=\tanh \left(W_{f, l} * \mathcal{X}_{out}^{l-1}\right) \odot \sigma\left(W_{g, l} * \mathcal{X}_{out}^{l-1}\right)
\end{equation}

where $\mathcal{X}_{out}^{i} \in \mathbf{R}^{C^{i} \times N \times T^{i}}$ is the output of the $i^{th}$ ST-block, $C^{i}$ and $T^{i}$ are the quantity of channels of the input ST data in the $i^{th}$ ST-block, and the size of the temporal dimension in the $i^{th}$ ST-block, respectively. $\sigma$ denotes the sigmoid activation function, $\odot$ represents an element-wise multiplication operator, $*$ represents a convolution operator, $f$ and $g$ mean filter and gate, and $W$ is the learnable convolution filter.

Received inputs from the final layer as $[C^{l-1},N, T^{l-1}]$-sized three-dimension tensors.


After going through the gated TCN block, the outputs of it turn into three-dimension tensors with the size of $[C^{l}, N, T^{l}]$ with $T^{l}=T^{l-1}-d^{l}$, where $d^{l}$ is the dilation size in the $l^{th}$ ST-block. As a result, the length of the temporal dimensions becomes shorter. In terms of the first ST-block, $\mathcal{X}_{out}^{0} = \operatorname{Conv}_{1*1}(\mathcal{X} \| \mathcal{M})\in \mathbf{R}^{C^{0} \times N \times T'}$, where ${Conv}_{1*1}$ is a 1*1 convolution network which is used to increase dimensions, and $\|$ symbolizes the concatenation operation. 

\subsection{Attention structure for extracting dynamic spatial information}


\begin{figure}[htbp]
\centering
\includegraphics[width=0.5\textwidth]{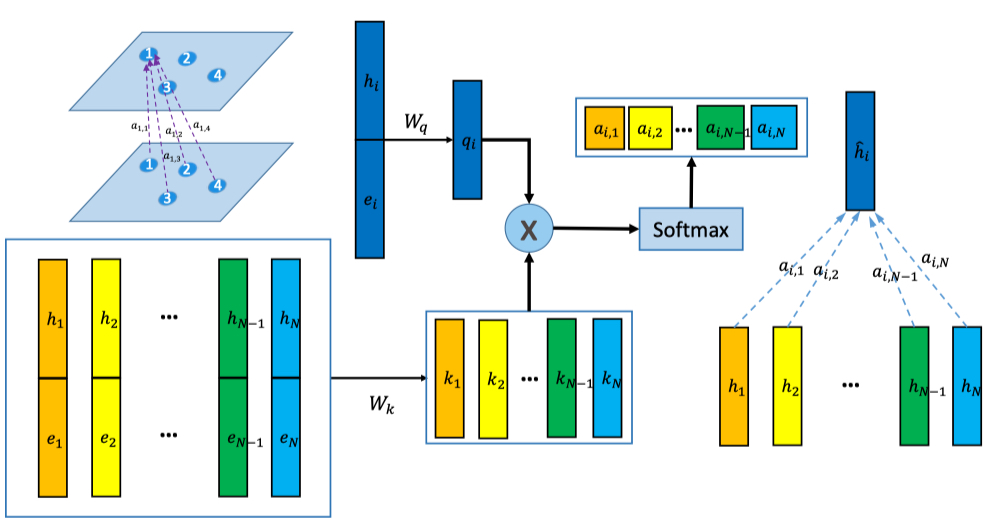}
\caption{Framework of {DAN}}
\label{attention}
\end{figure}

The essential concept of attention is to dynamically assign matching weights to various nodes from both the past and the future, as shown in Fig \ref{attention}. For node $i$, the attention structure allows us to calculate a weighted sum of information from all the information of remaining nodes in the network:
\begin{equation}
    {h'}^{l}_{i,t}=\sum_{j \leq N} \alpha_{i, j} \cdot {h}^{l}_{j,t},
    \label{att_eq}
\end{equation}
where $\alpha_{i, j}$ is the attention weight indicating the significance of node $j$ to node $i$ with $\sum_{j \leq N} \alpha_{i, j}=1$, ${h}^{l}_{i,t}=\mathcal{X}_{T}^{l}[:,i,t]$ and ${h}^{l}_{i,t} \in \mathbf{R}^{C^{l}}$. Intuitively, a higher weight represents the higher importance of node $j$ to node $i$. 


Previous studies and practical traffic data research \cite{tian2021spatial, pan2013short} have shown that the ST data prediction of future conditions can benefit from a carefully designed traffic network structure. We can naturally assume that the ST data imputation can also benefit from it. Motivated by this intuition, we integrate not only the intrinsic network information but also the traffic situation information into the novel imputation model.
After concatenating the learnable node embedding with the hidden state representation, and in order to understand the hidden representations of each node in the network we adopt a novel structure and learn a node embedding. Regarding the previous complex factors that affect the relationship between nodes at every time step, we adopt a new structure to learn a node embedding to grasp the hidden representations of all nodes in the network. Node embedding is helpful in deep neural networks as it can transfer the discrete variables to continuous variables and reduce the dimensionality of categorical variables, as well as meaningfully represent categories in the new space. Following, we compute

\begin{equation}
    e_{i, j}=\frac{\langle W_q ({h}^{l}_{i,t} \| e_i), W_k ({h}^{l}_{j,t} \| e_j) \rangle}{\sqrt{d'}},
\end{equation}
where $\|$ denotes the concatenation operator, $\langle \cdot,\cdot \rangle$ indicates the inner product operation, $e_i$ is the node embedding representation of the node $i$, $W_k, W_q \in \mathbb{R}^{d' \times d_c}$ are the key and query matrix in {DAN}, respectively, and $d_c$ is the dimension of ${h}^{l}_{i,t} \| e_i$. The node embedding $e_i$, the key matrix $W_k$, and query matrix $W_q$ are all learnable parameters from neural networks.

\begin{equation}
    \alpha_{i, j}=\frac{\exp (e_{i, j})}{\sum_{k \leq N} \exp \left(e_{i, k}\right)}
\end{equation}

After computing the attention scores, the parameters of hidden states will be updated by Equation \ref{att_eq}. The output dimension is identical to input with $\mathcal{X}_{S}^{l}[:,i,t] = {h'}^{l}_{i,t}$. This operation is useful as we do not introduce any non-parameter information and the computing process of node pairs is parallelizable in the training process.

\subsection{The output layer}
ResNet architecture is used in our structure to deliver information into deeper neural networks and prevent gradient problems. The output of the $l^{th}$ ST-block gains by:

\begin{equation}
    \mathcal{X}_{out}^{l}=\mathcal{X}_{S}^{l} + \mathcal{X}_{out}^{l-1}.
\end{equation}


For every ST-block, the gated TCN is equipped with a skip output, shown in Fig \ref{stawnet}. Then, the skip connections are all added up by
$
    \sum_{i \leq N_{st}} {Conv}_{1*1}(\mathcal{X}_{T}^{i}),
$
where $N_{st}$ is the quantity of ST-blocks. Following, the neural network layers of a linear transformation with ReLU activation function are applied to calculate the final output. It represents the imputation ST data that are missing in the training dataset. During the training process, the object is to minimize the error between the ground-truth observations of traffic on the roads and the outputs from neural networks. The object function is:
\begin{equation}
    \operatorname{Loss}=(X^t - \hat{X}^t)^2, 
    \label{loss}
\end{equation}
where $t$ indicates the missing ST data, and $X^t$ and $\hat{X}^t$ denote the ground truth ST data and the output from the imputation model, respectively.

\section{Result and discussion}

We use two models as our benchmark: Bayesian Temporal Matrix Factorization (BTMF) \cite{BTFC} and Low-Rank Tensor Completion with Truncated Nuclear Norm minimization (LRTC-TNN) \cite{Zhou2018}.

The vector auto-regressive (VAR) model is integrated into the latent temporal components of the fully Bayesian matrix factorization model known as the BTMF. When performing imputation tasks, BTMF outperforms conventional matrix factorization models (without temporal modeling) and tensor factorization models in terms of accuracy thanks to the flexible VAR process. Following the author's implementation, we kept the same hyper-parameters: burn-in iterations of 1100, number of samples of 200, and rank depending on the data set (10 for the GZ, 50 for the HZ, and 30 for PeMS).

Low-rank tensor completion (LRTC) is used to modify the missing data imputation problem in spatiotemporal traffic data and define a new truncated nuclear norm (TNN). The parameters used for the model were: alpha of $(\frac{1}{3}, \frac{1}{3}, \frac{1}{3})$, $\rho$ of $e^{-5}$ , $\theta$ of 0.30, epsilon of $e^{-4}$ , and maximum iterations of 200.

For all the data sets, we performed a random input of missing data with rates of 20\%, 40\%, 60\%. However, because of the limitation of the paper, we just list 20\%, 40\%, 60\% holds the similarity between those two.


\begin{table}[htbp]
\centering
\caption{Experiments of the three models}

\begin{tabular}{|c|c|ccc|cc|} 
\hline
\multirow{2}{*}{Dataset} & \multirow{2}{*}{Model} & \multicolumn{3}{c|}{20\%}                       & \multicolumn{2}{c|}{40\%}       \\ 
\cline{3-7}
                         &                        & MAE            & MAPE          & RMSE           & MAE            & MAPE           \\ 
\hline
                         & BTMF                   & 2.80           & 0.10          & 4.23           & 2.82           & 0.10           \\
GZ                       & LRTC-TNN               & 1.99           & 0.07          & 2.88           & 2.15           & 0.07           \\
                         & STAWnet                & \textbf{1.73}  & \textbf{0.06} & \textbf{2.49}  & \textbf{1.85}  & \textbf{0.06}  \\ 
\hline
                         & BTMF                   & 17.90          & 0.22          & 42,34          & 19.09          & 0.25           \\
HZ                       & LRTC-TNN               & \textbf{14.83} & \textbf{0.19} & \textbf{25.99} & \textbf{15.20} & \textbf{0.20}  \\
                         & STAWnet                & 18.26          & 0.29          & 32.54          & 19.55          & 0.29           \\ 
\hline
                         & BTMF                   & 60.01          & 1.01          & 60.55          & 58.93          & 1.00           \\
PeMS                     & LRTC-TNN               & 58.89          & 1.00          & 60.41          & 58.89          & 1.00           \\
                         & STAWnet                & \textbf{3.36}  & \textbf{0.08} & \textbf{4.53}  & \textbf{3.52}  & \textbf{0.08}  \\
\hline
\end{tabular}
\label{table2}
\end{table}

From Table \ref{table2}, We can observe that our model outperform the benchmark models in two data sets out of three. For the HZ data set, we observe that its size is relatively smaller than the other two, and this probably limited the capability of our neural network model to create good representations of the data.

To evaluate this assumption, we performed a sensitivity analysis on the PeMS data, running the same model for different portions of the data: 50\%, 80\%, and 100\%, and compare the performance on each scenario. The result of the sensitivity analysis is displayed on Table \ref{table3}.
\begin{table}[htbp]
\centering
\caption{Sensitivity analysis on the size of data}
\begin{tabular}{|l|l|l|l|} 
\hline
\%data & MAE & MAPE & RMSE  \\ 
\hline
50\%   & 0.1 & 0.1  & 0.1   \\
70\%   & 0.2 & 0.2  & 0.2   \\
90\%   & 0.3 & 0.3  & 0.3   \\
100\%  & 0.4 & 0.4  & 0.4   \\
\hline
\end{tabular}
\label{table3}
\end{table}

As we can see from Table \ref{table3}, the performance can be undermined when the amount of data is not enough for the neural network to learn the representation properly.

Apart from this setback, we believe that the proposed model can contribute to the analysis of trafﬁc when the available historical data is reasonable, as suggested by the results obtained on the GZ and PeMS data sets.

\section{Conclusion and future work}

Due to the continuous changes in traffic conditions on the road network, ST data imputation is highly challenging. It does, however, display links and dependencies that are both spatial and temporal, and we may effectively use them to extract correlations to improve performance in ST data imputation tasks. In the current work, we present a new approach to effectively collect spatial-temporal information by fusing attention structures and temporal convolution networks. Our algorithm outperforms previous methods in three real-world datasets with a large margin. Additionally, the interpretability of our suggested model is partially demonstrated by the learnable node embedding and attention weights, which can assist in determining the significance of its past and future nodes.

\bibliographystyle{IEEEtran}  
\bibliography{main}

\end{document}